%% file: main.tex
\documentclass[sigconf]{acmart}

\usepackage{graphicx}
\usepackage{url} 
\usepackage{amsmath}
\usepackage{amsthm}
\usepackage{amsfonts}
\usepackage{bm}
\usepackage{caption}
\usepackage{adjustbox}
\usepackage{multirow}
\usepackage[english]{babel}
\usepackage{subfigure}
\usepackage{booktabs}
\usepackage{array}
\usepackage{xcolor} 
\usepackage{pifont}
\usepackage{algorithmic}
\usepackage[linesnumbered,ruled,vlined]{algorithm2e}
\usepackage[most]{tcolorbox}
\usepackage{enumitem}

\usepackage{mathtools}

\usepackage{amsmath} 
\newcommand{\Eq}[1]{Eq.~\eqref{#1}}
\newcommand{\Eqs}[2]{Eq.~\eqref{#1}--\eqref{#2}}

\theoremstyle{plain}

\theoremstyle{definition}

\theoremstyle{remark}

\def\model{scFM}

\AtBeginDocument{%
  }

\setcopyright{acmlicensed}
\copyrightyear{2026}
\acmYear{2026}
\acmDOI{XX}
\acmConference[Conference acronym 'XX]{xx}{2026}{xx}

\begin{document}

\title{From Snapshots to Trajectories: Learning Single-Cell Gene Expression Dynamics via Conditional Flow Matching}

\author{Siyu Pu\textsuperscript{1}, Qingqing Long\textsuperscript{1}, Xiaohan Huang\textsuperscript{1}, Haotian Chen\textsuperscript{1}, Jiajia Wang\textsuperscript{1}, \\Meng Xiao\textsuperscript{1}, Xiao Luo\textsuperscript{2}, Hengshu Zhu\textsuperscript{1}, Yuanchun Zhou\textsuperscript{1}, Xuezhi Wang\textsuperscript{1}}
\affiliation{%
  \institution{\textsuperscript{1}Computer Network Information Center, Chinese Academy of Sciences}
  \institution{\textsuperscript{2}University of California, Los Angeles}
   \country{}
}

\renewcommand{\shortauthors}{Siyu Pu et al.}

\begin{abstract}
Single-cell RNA sequencing (scRNA-seq) provides high-dimensional profiles of cellular states, enabling data-driven modeling of cellular dynamics over time. 
In practice, time-resolved scRNA-seq is collected at only a few discrete time points as unpaired snapshot populations, leaving substantial temporal gaps.  
This motivates trajectory inference at unmeasured time points. 
Existing methods mainly follow two directions, optimal-transport (OT) alignment provides distribution-level matching between observed snapshots, while continuous-time generative models support forecasting via learned dynamics. 
However, two challenges remain: 
(i) unpaired snapshots render local transitions between adjacent time points ambiguous, leading to unstable supervision; 
and (ii) long-horizon prediction relies on repeated integration, where small modeling errors compound and cause distribution drift. 
To address these challenges, we propose single-cell Flow Matching (\textbf{\model}), a latent generative framework based on coupling-conditioned flow matching. 
First, we compute entropically regularized OT couplings between adjacent snapshots and use them to construct soft, weighted flow-matching targets for learning time-dependent velocity fields. 
Second, we learn bidirectional velocity fields and leverage their consistency to refine couplings and improve temporal coherence under sparse supervision. 
Third, we introduce distribution-level alignment and latent dynamic regularization to anchor long rollouts and mitigate drift. 
Experiments on real-world time-series scRNA-seq datasets show that \model consistently improves distributional prediction performance for both temporal interpolation and extrapolation. 
Moreover, \model yields more accurate trajectory reconstruction and temporally coherent visualizations where intermediate time points are absent, indicating a more faithful recovery of underlying temporal gene expression dynamics.
 
\end{abstract}


\maketitle

\input{chapter/0_intro}

\input{chapter/1_related}
\input{chapter/2_0_pre}
\input{chapter/2_model}

\input{chapter/3_exp}
\input{chapter/4_conclusion}

\bibliographystyle{ACM-Reference-Format}
\bibliography{cite}

\end{document}

%% file: chapter/0_intro.tex
\section{Introduction}
Single-cell RNA sequencing (scRNA-seq) has transformed the study of complex biological systems by profiling gene expression at single-cell resolution, thereby revealing cellular heterogeneity and state transitions that are invisible to bulk measurements~\cite{guo2025computational,marx2024scrna,saliba2014single,haque2017practical}.
Modeling the resulting continuous gene-expression dynamics is essential for understanding time-varying cellular programs and critical transition phases.
However, time-resolved scRNA-seq \textit{in vivo} remains challenging: sequencing is destructive and thus provides \emph{unpaired} snapshot distributions across time rather than longitudinal trajectories, while dense sampling is often infeasible, leaving large temporal gaps and missing intermediate states~\cite{ding2022temporal,yin2024dyna,zhang2025benchmarking,erhard2022time,schiebinger2019optimal,zhang2024learning,golumbeanu2018single}.
Consequently, a core computational goal is to generate \emph{in silico} cells at unobserved time points to recover continuous dynamics from sparsely sampled snapshots~\cite{guo2025computational,sha2024reconstructing,la2018rna}.

\begin{figure}[htbp]%
    \centering
    \includegraphics[width=\linewidth]{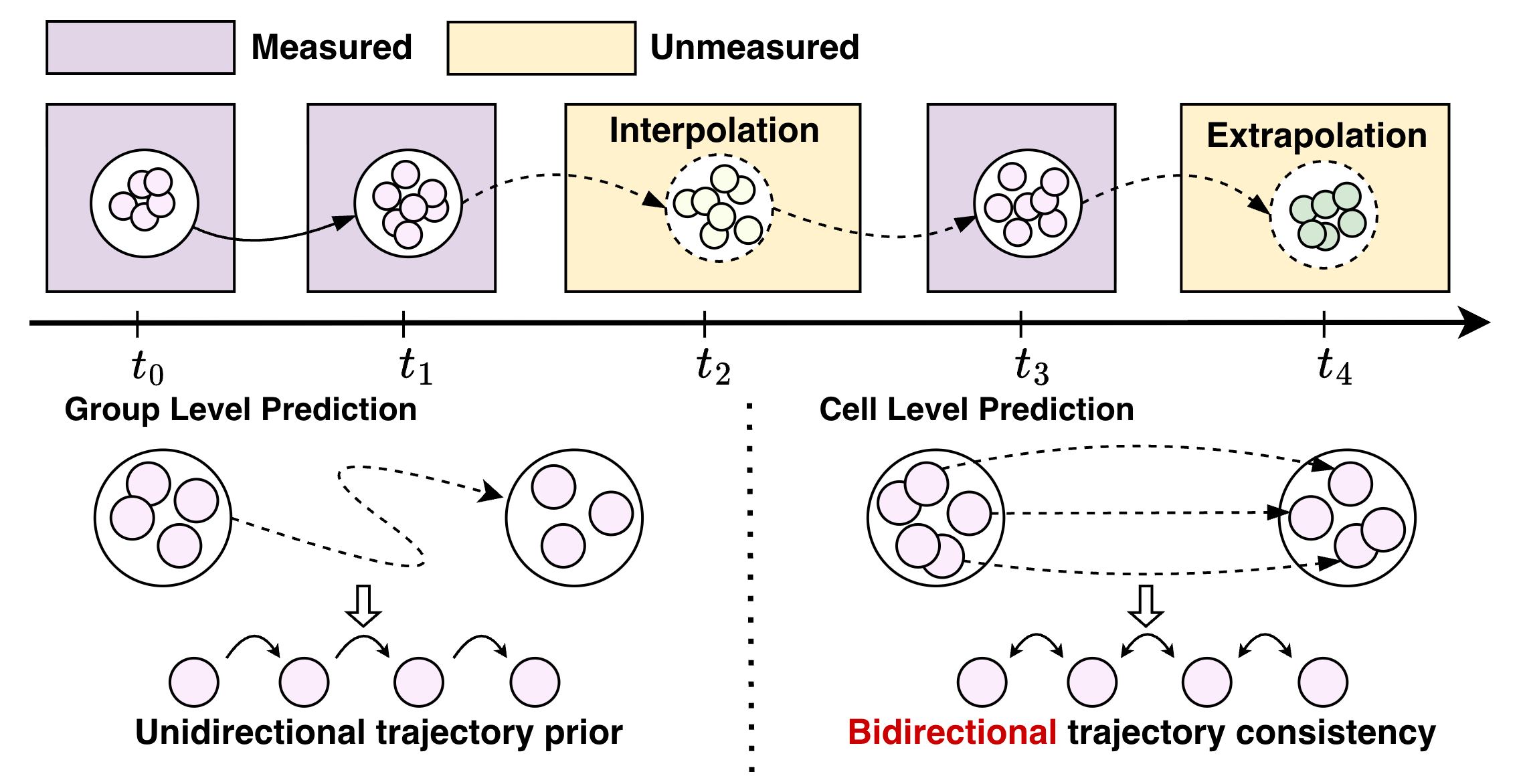}
    \caption{Given sparsely sampled scRNA-seq snapshots, \model{} integrates OT-informed alignment and simulation-free flow matching to learn time-conditioned single-cell dynamics and generate \emph{in silico} cells at unobserved time points.}
    \label{fig:overview}
    \vspace{-2mm}
\end{figure}

A spectrum of computational approaches has been explored in recent years~\cite{sha2024reconstructing,bergen2020generalizing,huguet2022manifold,tong2020trajectorynet,yeo2021generative,zhang2024scnode}.
Trajectory inference and pseudotime methods recover putative developmental orderings but do not directly model physical time or provide a generative mechanism for predicting future snapshots~\cite{trapnell2014dynamics,haghverdi2016diffusion,street2018cell,saelens2019comparison}.
RNA-velocity-based methods estimate local directions of state changes, yet rely on additional assumptions and typically provide local, short-horizon guidance rather than distribution-level long-range prediction~\cite{bergen2020generalizing,lange2022cellrank,gayoso2022deep}.
Optimal-transport (OT) approaches align snapshot distributions across time and have shown strong performance in reconstructing ancestor-descendant relations~\cite{schiebinger2019optimal,klein2025mapping}, while extensions further incorporate population growth or unbalanced mass transport~\cite{sha2024reconstructing}.
In parallel, continuous-time generative dynamics models (e.g., neural ODE/flow-based models) attempt to learn time-conditioned vector fields from snapshots for interpolation or forecasting~\cite{yeo2021generative,tong2020trajectorynet,huguet2022manifold,zhang2024scnode}. However, existing methods still face \textbf{several limitations} for embryonic scRNA-seq:

\textbf{(1) Ambiguous cell-level temporal dependencies under unpaired supervision.}
Snapshot measurements provide unordered populations at each time, so the learning signal is inherently \emph{set-valued}: there is no ground-truth correspondence that specifies which micro-transitions between adjacent snapshots are plausible.
Many existing objectives therefore default to coarse distribution matching or one-directional constraints, which can leave the local transition structure underdetermined.
This ambiguity is especially problematic when time gaps are large or trajectories branch/merge, where multiple distinct cell-level transports can explain the same marginal snapshots, yielding weakly identifiable dynamics and temporally incoherent rollouts.

\textbf{(2) Loose coupling between snapshot alignment and vector-field learning.}
\sloppy
OT-based alignment provides strong distribution-level matching, while dynamics models aim to learn a time-conditioned vector field whose integration reproduces snapshot marginals.
However, in many pipelines these two components are only weakly connected: OT is used as a post-hoc alignment, and the vector field is optimized with objectives that do not directly inherit the OT-induced correspondence structure.
As a result, the supervision provided to the vector field can be unstable or noisy, especially when snapshots are far apart-because the model must simultaneously infer pairings and fit dynamics without a coupling-aware training target.

\textbf{(3) Compounding drift in long-horizon generation without global anchors.}
Continuous-time models typically generate cells at future times by repeatedly integrating a learned vector field.
Even small local regression errors can accumulate over time, causing the predicted population to gradually drift away from the true snapshot manifold and violate observed marginals.
Local supervision alone is often insufficient to prevent such error compounding, because it does not explicitly penalize distribution mismatch at distant time points.

To address these issues, we propose Single-Cell Flow Matching (\textbf{\model{}}), a single-cell generative framework for developmental dynamics modeling from sparse scRNA-seq snapshots.
At a high level, \model{} combines OT-informed alignment with simulation-free flow matching~\cite{lipman2022flow,pooladian2023multisample} to learn time-conditioned vector fields that can generate realistic single-cell distributions at unmeasured time points.
Crucially, \model{} explicitly leverages \emph{directional} temporal dependencies between successive snapshots and integrates information from both preceding and subsequent states, providing more constrained and informative supervision for learning developmental dynamics.
To further stabilize long-horizon generation, \model{} incorporates a global distribution-level anchoring mechanism that regularizes the learned dynamics to reduce drift away from observed snapshot distributions.
By operating at the cell level, \model{} preserves cellular heterogeneity and supports fine-grained trajectory reconstruction in embryogenesis.
We summarize the main contributions of this work as follows:
\begin{itemize}[leftmargin=*, itemsep=0pt, topsep=2pt]
    \item We learn forward and backward time-conditioned velocity fields and enforce their consistency, enabling the model to exploit information from both preceding and subsequent snapshots to better constrain local transitions under unpaired supervision.
    \item We tightly couple snapshot alignment and vector-field learning by integrating entropically regularized OT couplings into simulation-free flow matching, converting distribution-level alignment into soft, weighted cell-level supervision for time-series prediction.
    \item We introduce distribution-level regularization that anchors long rollouts to observed snapshot marginals, reducing compounding drift and improving robustness in long-horizon extrapolation.
    \item We conduct extensive experiments on scRNA-seq datasets, demonstrating consistent improvements over SOTA baselines in dynamics learning.
\end{itemize}

%% file: chapter/1_related.tex
\section{Related Work}
We organize prior work on modeling time-resolved scRNA-seq snapshots into three paths: OT-based couplings for aligning unpaired populations, continuous-time generative dynamics, and simulation-free regression objectives.

\subsection{OT for Snapshot Alignment}
Optimal transport (OT) offers a principled mechanism for aligning cell populations across time by estimating couplings that minimize a transportation cost \cite{bunne2024optimal,schiebinger2019optimal}. Practical OT pipelines frequently rely on entropic regularization and Sinkhorn-type objectives for scalability, and Sinkhorn divergences provide a statistically better-behaved alternative for comparing distributions \cite{cuturi2013sinkhorn,feydy2019interpolating}. For developmental systems with proliferation/death, unbalanced OT and entropy-transport formulations (e.g., Hellinger-Kantorovich) provide a natural way to relax mass conservation \cite{chizat2018unbalanced,liero2018optimal}, and the computational OT literature supplies the algorithmic foundations \cite{peyre2019computational}. In the context of temporal prediction, OT-based methods are often used as baselines or auxiliary supervision: couplings between observed time points can guide interpolation, while genuine forecasting beyond the last observation is more challenging because no target marginal is available to anchor the coupling.Overall, OT-based alignment can recover distribution-level correspondences and support short-range interpolation, but by itself it lacks an explicit time-conditioned dynamics model and provides limited guidance for long-horizon generation under sparse observations.

\subsection{Continuous-Time Generative Dynamics}
A complementary line of work learns continuous-time generative dynamics that transport an initial distribution to later snapshot marginals by integrating a learned vector field~\cite{grathwohl2018ffjord,fang2021spatial}.
TrajectoryNet \cite{tong2020trajectorynet} connects continuous normalizing flows with dynamic OT regularization to control paths between snapshot distributions, enabling non-linear transport trajectories in continuous time.
MIOFlow\cite{huguet2022manifold} further incorporates manifold-aware geometry by operating in an autoencoder latent space and penalizing OT discrepancies with manifold-sensitive ground costs, improving interpolation fidelity when data lie on complex manifolds.
PRESCIENT\cite{yeo2021generative} models differentiation using an energy landscape and enables simulation under interventions, emphasizing mechanistic structure alongside prediction.
Closer to the explicit forecasting objective, scNODE couples a VAE with latent neural ODE ~\cite{chen2018neural} dynamics and regularization to improve robustness when generating cells at unobserved time points, making it a strong and widely used baseline for time-point prediction.In summary, continuous-time models can represent smooth time-conditioned transport and enable forecasting via ODE integration, but learning accurate vector fields from unpaired snapshots often requires strong and stable supervision; without careful coupling between alignment and dynamics, local ambiguity and long-horizon drift can still degrade predictions.

\subsection{Simulation-Free Flow Objectives}
Recent advances in generative modeling have introduced simulation-free \emph{regression-style} objectives for learning continuous-time flows~\cite{ho2020denoising}.
Flow matching\cite{lipman2022flow} trains a vector field by regressing to analytically defined probability paths, and conditional/OT-informed variants improve stability and generalization by leveraging coupling-aware paths, including minibatch OT constructions\cite{tong2023improving}.
These objectives have motivated time-series imputation and missing-time-point modeling beyond biology, e.g., multi-marginal flow matching formulations for dynamics across time and conditions\cite{rohbeck2025modeling}.
In the single-cell domain, DeepRUOT\cite{zhang2024learning} focuses on recovering snapshot-consistent stochastic dynamics under its transport-based formulation. CellStream\cite{ling2025cellstream} learns dynamics-informed embeddings that better represent temporal structure and improve robustness in downstream trajectory reconstruction. VGFM~\cite{wang2025joint} applies flow matching to snapshot dynamics and is often reported under hold-one-out evaluation protocols.
Importantly, this protocol reflects an evaluation choice rather than a hard modeling constraint: VGFM learns a continuous-time dynamics that can be queried at arbitrary times after training, and thus can be used to generate predictions for multiple held-out timepoints within a single trained model under our multi-holdout setting.
 Overall, these works provide valuable inductive biases for temporal structure; in contrast, our emphasis is on improving distribution-level time-point prediction accuracy under multi-timepoint holdout protocols that more closely reflect realistic experimental gaps.
Overall, simulation-free objectives can provide stable and scalable velocity-field learning, but existing single-cell methods may still rely on one-directional or loosely coupled alignment signals, leaving cell-level temporal dependencies underconstrained and long-horizon rollouts susceptible to drift.
Building on this line, our \model{} tightly couples OT-derived soft correspondences with flow-matching regression (yielding coupling-conditioned supervision), further strengthens identifiability via bidirectional velocity fields, and introduces global distribution-level anchoring to stabilize long-horizon generation.

%% file: chapter/2_0_pre.tex
\section{Preliminaries}

\paragraph{Problem Definition} We observe temporally resolved single-cell transcriptomic snapshots.
Let $t \in \mathcal{T}$ denote a physical time index.
At each observed time $t$, we are given an unordered cell set
$\mathcal{X}_t = \{x_i^t\}_{i=1}^{n_t} \subset \mathbb{R}^{G}$,
where $G$ is the number of genes and each $x_i^t$ is a vector of normalized expression values.
We write $p_t$ for the unknown data distribution that generates $\mathcal{X}_t$.
Let $\mathcal{T}_{\mathrm{tr}} = \{t_0 < t_1 < \cdots < t_M\}$ be the training timepoints.
Our goal is to learn a continuous-time generative dynamics model that, given cells at the earliest time $t_0$,
can generate cell populations at any $t \in [t_0, t_M]$,
and can extrapolate beyond $t_M$ when the learned dynamics generalizes.

\paragraph{Latent representation via a shared VAE}
Direct modeling in gene space is challenging due to high dimensionality and sparsity. Followed by exist work~\cite{zhang2024scnode}, we first train a VAE~\cite{kingma2013auto} which can capture complex cellular relationships and variations in single-cell data.
We then model cellular development as a continuous-time dynamical flow in the latent space. The encoder $\mathrm{Enc}_{\phi}$ maps gene expression to a Gaussian latent distribution.
For each cell $x \in \mathbb{R}^{G}$,
\begin{equation}
\label{eq:vae-enc}
\begin{aligned}
(\mu_{\phi}(x), \sigma_{\phi}(x)) &= \mathrm{Enc}_{\phi}(x), \\
q_{\phi}(z \mid x) &= \mathcal{N}\left(z \mid \mu_{\phi}(x),
\operatorname{diag}\bigl(\sigma_{\phi}(x)^2\bigr)\right).
\end{aligned}
\end{equation}
We draw a reparameterized sample
\begin{equation}
z = \mu_{\phi}(x) + \sigma_{\phi}(x) \odot \varepsilon,
\qquad
\varepsilon \sim \mathcal{N}(0, I_d).
\label{eq:reparam}
\end{equation}
The decoder $\mathrm{Dec}_{\theta}$ maps latents back to gene space,
\begin{equation}
\hat{x} = \mathrm{Dec}_{\theta}(z).
\label{eq:vae-dec}
\end{equation}
Then we train the VAE using a reconstruction loss and a KL regularizer.
For a mini-batch $\{x_b\}_{b=1}^{B}$,
\begin{equation}
\label{eq:vae-loss}
\begin{split}
\mathcal{L}_{\mathrm{VAE}}(\phi,\theta)
&=
\frac{1}{B}\sum_{b=1}^{B}\bigl\lVert x_b - \mathrm{Dec}_{\theta}(z_b)\bigr\rVert_2^2 \\
&\quad +
\lambda_{\mathrm{KL}}\,
\frac{1}{B}\sum_{b=1}^{B}
\mathrm{KL}\left(q_{\phi}(z\mid x_b)\,\|\,\mathcal{N}(0,I_d)\right).
\end{split}
\end{equation}

%% file: chapter/2_model.tex
\section{Model: \model}
\label{sec:model}
\begin{figure*}[htbp]%
    \centering { \includegraphics[width=0.9\linewidth]{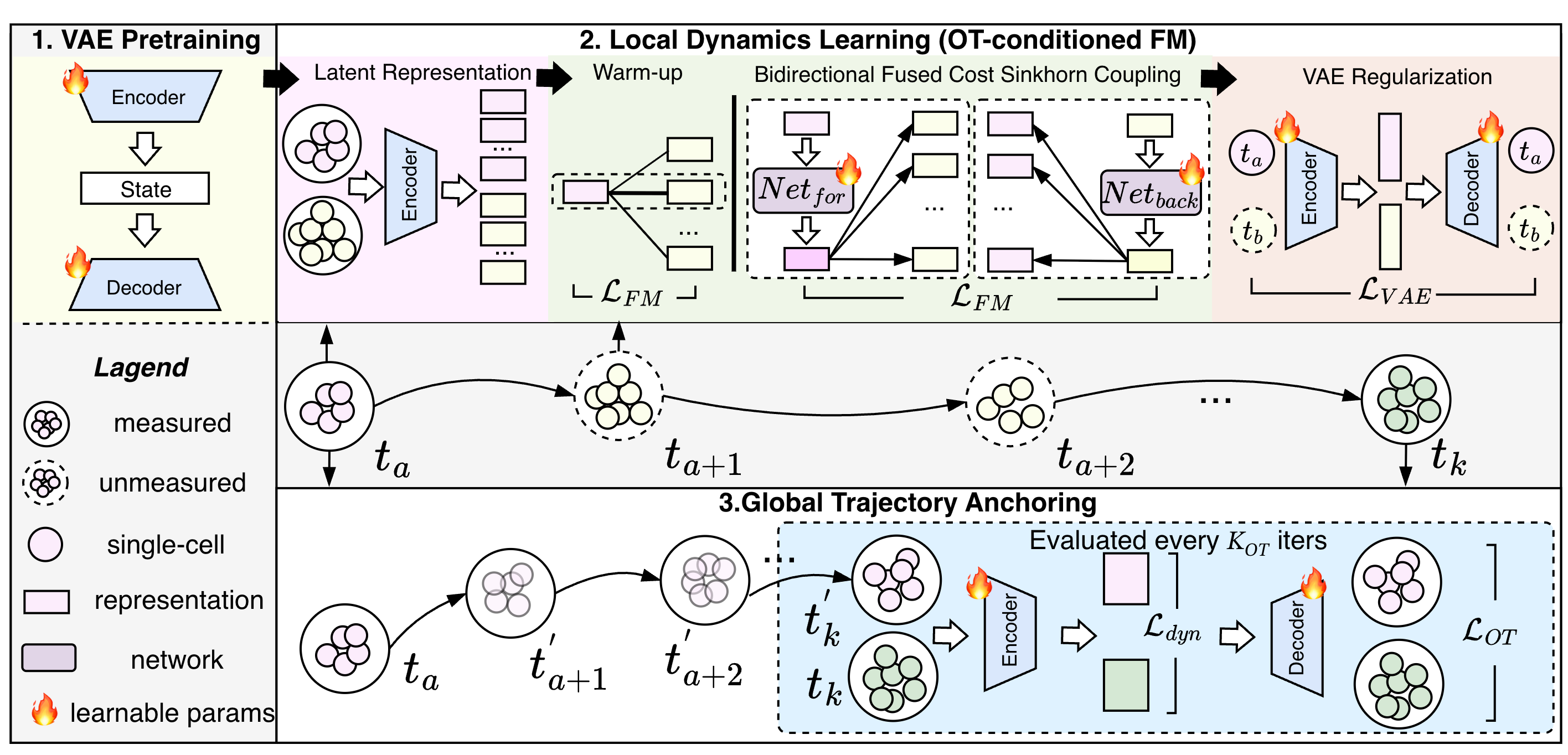} }
    \vspace{-3mm}
    \caption{scFM models continuous-time single-cell dynamics in a VAE latent space using a two-stage training framework. 
    (1) A shared VAE is first pretrained to learn compact gene-expression representations. 
    (2) Local OT-conditioned flow matching learns time-dependent velocity fields between adjacent snapshots using entropic optimal transport with a bidirectional fused cost and symmetric supervision. 
    (3) During global trajectory anchoring, representation regularization and periodic global OT anchoring enforce reconstruction fidelity and long-horizon dynamic consistency.
    }
    \label{fig:framework}
    \vspace{-2mm}
\end{figure*}

In this paper, we propose \textbf{s}ingle-\textbf{c}ell \textbf{F}low \textbf{M}atching (\model), a latent generative framework based on coupling-conditioned flow matching.
The framework of \model~is shown in Fig.\ref{fig:framework}. First, we compute entropically regularized OT couplings between adjacent snapshots and use them to construct soft, weighted flow-matching targets for learning time-dependent velocity fields. Second, we learn bidirectional velocity fields and leverage their consistency to refine couplings and improve temporal coherence under sparse supervision. Third, we introduce distribution-level alignment and latent dynamic regularization to anchor long rollouts and mitigate drift.

\subsection{Continuous-time latent dynamics with two velocity fields}
Single-cell data are observed as \emph{unpaired} snapshots across time, where cell-wise correspondences are unknown.
Inspired by entropic OT for matching distributions across snapshots~\cite{cuturi2013sinkhorn,peyre2019computational}, we estimate a soft coupling between adjacent time points and use it to supervise latent dynamics.
While a single forward model can generate future trajectories, the lack of correspondence makes the coupling and velocity supervision highly ambiguous and prone to error accumulation; we therefore learn \emph{bidirectional} time-dependent velocity fields (forward and backward) to impose a symmetric constraint that stabilizes coupling construction and velocity learning.

We introduce a forward field $v_{\psi}^{\mathrm{f}} : [t_0,t_M]\times \mathbb{R}^{d}\to\mathbb{R}^{d}$
and a backward field $v_{\psi}^{\mathrm{b}} : [t_0,t_M]\times \mathbb{R}^{d}\to\mathbb{R}^{d}$.
The forward latent trajectory $z^{\mathrm{f}}(t)$ solves
\begin{equation}
\frac{d z^{\mathrm{f}}(t)}{dt} = v_{\psi}^{\mathrm{f}}\bigl(t, z^{\mathrm{f}}(t)\bigr),
\qquad
z^{\mathrm{f}}(t_0) = z_0.
\label{eq:fwd-ode}
\end{equation}
The backward latent trajectory $z^{\mathrm{b}}(t)$ solves
\begin{equation}
\frac{d z^{\mathrm{b}}(t)}{dt} = v_{\psi}^{\mathrm{b}}\bigl(t, z^{\mathrm{b}}(t)\bigr),
\qquad
z^{\mathrm{b}}(t_1) = z_1.
\label{eq:bwd-ode}
\end{equation}
At inference time, we sample $z_0 \sim q_{\phi}(z \mid x^{t_0})$ from observed cells at $t_0$,
integrate Eq.~\eqref{eq:fwd-ode} to obtain $z^{\mathrm{f}}(t)$ at desired times,
then decode $\hat{x}(t)=\mathrm{Dec}_{\theta}(z^{\mathrm{f}}(t))$.
The backward field is used during training to refine coupling construction and to provide symmetric supervision.

\subsection{OT-conditioned flow matching for local supervision}
A key difficulty is that snapshots are unordered.
There is no ground-truth correspondence between individual cells at adjacent timepoints.
We therefore construct a soft coupling between mini-batches at adjacent times using optimal transport,
and we use that coupling as a supervision distribution for flow matching.
This choice avoids committing to hard matchings and yields a principled distribution-level pairing that is robust to sampling noise.

Let $\mathcal{E}_{\mathrm{tr}} = \{(t_k, t_{k+1})\}_{k=0}^{M-1}$ denote the set of adjacent intervals. First, we choose adjacent times $(t_a,t_b)\in\mathcal{E}_{\mathrm{tr}}$ with $t_b>t_a$ and sample two mini-batches
$\{x_i^{a}\}_{i=1}^{B}\subset\mathcal{X}_{t_a}$ and $\{x_j^{b}\}_{j=1}^{B}\subset\mathcal{X}_{t_b}$.
After encoding them into latents,
$z_i^{a}\sim q_{\phi}(z\mid x_i^{a})$ and $z_j^{b}\sim q_{\phi}(z\mid x_j^{b})$,
we define a pairwise cost matrix $C\in\mathbb{R}^{B\times B}$ for OT.
During an initial warmup stage, we adopt a geometric (squared Euclidean) cost:
\begin{equation}
C^{\mathrm{euc}}_{ij} = \bigl\lVert z_i^{a} - z_j^{b} \bigr\rVert_2^2.
\label{eq:cost-euc}
\end{equation}

After warmup we use a bidirectional fused cost that incorporates both velocity fields.
Let $\Delta t = t_b - t_a$.
Using one-step Euler predictions,
\begin{equation}
\tilde{z}_{i}^{b} = z_i^{a} + v_{\psi}^{\mathrm{f}}(t_a, z_i^{a}) \cdot \Delta t,
\qquad
\tilde{z}_{j}^{a} = z_j^{b} + v_{\psi}^{\mathrm{b}}(t_b, z_j^{b}) \cdot (t_a - t_b).
\label{eq:euler-pred}
\end{equation}
We define the fused cost
\begin{equation}
C^{\mathrm{bi}}_{ij}
=
\frac{1}{2}\bigl\lVert \tilde{z}_{i}^{b} - z_j^{b} \bigr\rVert_2^2
+
\frac{1}{2}\bigl\lVert z_i^{a} - \tilde{z}_{j}^{a} \bigr\rVert_2^2.
\label{eq:cost-bi}
\end{equation}
This cost encourages consistency: a candidate pair is favored when forward dynamics maps $z_i^{a}$ close to $z_j^{b}$
and backward dynamics maps $z_j^{b}$ close to $z_i^{a}$.

\paragraph{Entropic OT coupling.}
Let $u = \frac{1}{B}\mathbf{1}$ be the uniform marginal.
We compute an entropic OT plan
\begin{equation}
\label{eq:sinkhorn}
\begin{split}
\pi
=
\arg\min_{\gamma \in \Pi(u,u)}
\ \langle \gamma, C \rangle
+
\varepsilon
\sum_{i=1}^{B}\sum_{j=1}^{B}
\gamma_{ij}\bigl(\log \gamma_{ij} - 1\bigr),
\end{split}
\end{equation}
where $\Pi(u,u)$ is the set of couplings with row and column sums equal to $u$.
We solve Eq.~\eqref{eq:sinkhorn} using Sinkhorn iterations~\cite{flamary2021pot}.
The resulting $\pi_{ij}$ defines a soft pairing distribution between the two unordered mini-batches.

\paragraph{OT-conditioned flow matching objective.}
Given a paired endpoint sample $(z_i^{a}, z_j^{b})$ drawn according to $\pi$,
we define a linear bridge trajectory between them.
Sample $\alpha \sim \mathrm{Unif}[0,1]$ and set
\begin{equation}
t_{\alpha} = (1-\alpha)t_a + \alpha t_b,
\qquad
z_{\alpha} = (1-\alpha)z_i^{a} + \alpha z_j^{b}.
\label{eq:bridge}
\end{equation}
The target constant velocity along this bridge is
\begin{equation}
u_{ij} = \frac{z_j^{b} - z_i^{a}}{t_b - t_a}.
\label{eq:target-vel}
\end{equation}
We train the forward velocity field by regression to this target velocity,
averaged over the OT coupling,
\begin{equation}
\label{eq:fm-fwd}
\mathcal{L}_{\mathrm{FM}}^{\mathrm{f}}(\psi)
=
\mathbb{E}_{(i,j)\sim \pi}
\mathbb{E}_{\alpha \sim \mathrm{Unif}[0,1]}
\bigl\lVert v_{\psi}^{\mathrm{f}}(t_{\alpha}, z_{\alpha}) - u_{ij} \bigr\rVert_2^2.
\end{equation}
We analogously define the backward flow matching loss by swapping endpoints,
\begin{equation}
\label{eq:fm-bwd}
\mathcal{L}_{\mathrm{FM}}^{\mathrm{b}}(\psi)
=
\mathbb{E}_{(i,j)\sim \pi}
\mathbb{E}_{\alpha \sim \mathrm{Unif}[0,1]}
\left\lVert v_{\psi}^{\mathrm{b}}(t_{\alpha}, z_{\alpha})
-
\frac{z_i^{a} - z_j^{b}}{t_a - t_b}
\right\rVert_2^2.
\end{equation}
The bidirectional local supervision is
\begin{equation}
\mathcal{L}_{\mathrm{FM}}(\psi)
=
\mathcal{L}_{\mathrm{FM}}^{\mathrm{f}}(\psi)
+
\mathcal{L}_{\mathrm{FM}}^{\mathrm{b}}(\psi).
\label{eq:fm-total}
\end{equation}

Eq.~(13)-(15) define the population objective averaging over all $B^2$ pairs under $\pi$.
In practice, we use a \emph{Top-$K$ approximation} by retaining only the $K$ largest coupling
weights per row. For each $i$, let $N_K(i)=\mathrm{TopK}(\pi_{i,\cdot})$,
$w_{ij}=\pi_{ij}\mathbf 1\{j\in N_K(i)\}$, and
$m=\sum_{i=1}^B\sum_{j\in N_K(i)} w_{ij}$.
We approximate the forward loss as
\begin{equation}
\label{eq:fm-topk-fwd}
\mathcal{L}^{(K),f}_{\mathrm{FM}}(\psi)
=\frac{1}{m}\sum_{i=1}^B\sum_{j\in N_K(i)} w_{ij}\;
\mathbb E_{\alpha\sim \mathrm{Unif}[0,1]}
\big[\|v^f_\psi(t_\alpha,z^{ij}_\alpha)-u_{ij}\|_2^2\big],
\end{equation}
and define $\mathcal{L}^{(K),b}_{\mathrm{FM}}$ analogously, with
$\mathcal{L}^{(K)}_{\mathrm{FM}}=\mathcal{L}^{(K),f}_{\mathrm{FM}}+\mathcal{L}^{(K),b}_{\mathrm{FM}}$.
This reduces the \emph{pairwise regression evaluations} from $O(B^2)$ to $O(BK)$
(the Sinkhorn coupling still uses the $B\times B$ cost matrix).
When $K\to B$ or the truncated mass $m$ is close to $1$, the approximation becomes tight;
for simplicity we denote $\mathcal{L}^{(K)}_{\mathrm{FM}}$ by $\mathcal{L}_{\mathrm{FM}}$ unless otherwise stated.

\subsection{Global distribution matching and latent dynamic regularization}
Local OT-conditioned flow matching provides informative supervision between adjacent snapshots,
but it does not directly constrain long-horizon rollouts.
We add global, distribution-level losses to (i) match predicted populations to observed snapshots and
(ii) regularize latent dynamics to stay consistent with the encoder geometry.

\paragraph{Global Wasserstein distribution matching.}
Let $\hat{\mathcal{X}}_{t_k}$ be predicted cells at time $t_k$ obtained by encoding cells at $t_0$,
rolling out Eq.~\eqref{eq:fwd-ode} to $t_k$, and decoding.
We measure the discrepancy between $\mathcal{X}_{t_k}$ and $\hat{\mathcal{X}}_{t_k}$ via an entropic OT cost
\begin{equation}
\label{eq:gene-ot}
\begin{split}
\mathrm{OT}\bigl(\mathcal{X}_{t_k}, \hat{\mathcal{X}}_{t_k}\bigr)
&=
\min_{\Gamma \in \Pi(u_k,\hat{u}_k)} \ \langle \Gamma, D_k \rangle
+
\varepsilon_k
\sum_{i=1}^{B}\sum_{j=1}^{B}
\Gamma_{ij}\bigl(\log \Gamma_{ij} - 1\bigr).
\end{split}
\end{equation}
where $D_k$ is a squared Euclidean cost matrix in gene space and $\Pi(u_k,\hat{u}_k)$ enforces marginal constraints.
We average this discrepancy over all training time points:
\begin{equation}
\mathcal{L}_{\mathrm{OT}}(\phi,\theta,\psi)
=
\frac{1}{M+1}\sum_{k=0}^{M}
\mathrm{OT}\bigl(\mathcal{X}_{t_k}, \hat{\mathcal{X}}_{t_k}\bigr).
\label{eq:ot-loss}
\end{equation}

\paragraph{Latent dynamic regularization.}
To prevent the learned dynamics from drifting away from the VAE latent geometry,
we align ODE-generated latents with encoder latents at each time point.
Let $\mathcal{Z}_{t_k}^{\mathrm{enc}}$ be a mini-batch of latents encoded from $\mathcal{X}_{t_k}$,
and let $\mathcal{Z}_{t_k}^{\mathrm{ode}}$ be the corresponding ODE-predicted latent set.
We define
\begin{equation}
\mathcal{L}_{\mathrm{dyn}}(\phi,\psi)
=
\frac{1}{M+1}\sum_{k=0}^{M}
\mathrm{OT}\bigl(\mathcal{Z}_{t_k}^{\mathrm{enc}}, \mathcal{Z}_{t_k}^{\mathrm{ode}}\bigr),
\label{eq:dyn-reg}
\end{equation}
where the OT cost is computed in latent space.

Finally, the overall \textbf{joint learning objective} combines representation learning, local flow matching, and the global constraints:
\begin{equation}
\label{eq:full-objective}
\begin{aligned}
\min_{\phi,\theta,\psi}\quad
&\mathcal{L}_{\mathrm{VAE}}(\phi,\theta)
+
\lambda_{\mathrm{FM}}\mathcal{L}_{\mathrm{FM}}(\psi) \\
&+
\lambda_{\mathrm{OT}}\mathcal{L}_{\mathrm{OT}}(\phi,\theta,\psi)
+
\lambda_{\mathrm{dyn}}\mathcal{L}_{\mathrm{dyn}}(\phi,\psi).
\end{aligned}
\end{equation}

Computing $\mathcal{L}_{\mathrm{OT}}$ and $\mathcal{L}_{\mathrm{dyn}}$ requires a long-horizon ODE rollout from $t_0$ to all $t_k\in\mathcal{T}_{\mathrm{tr}}$, which is substantially more expensive than the local per-pair objective.
We therefore use a simple schedule controlled by $K_{\mathrm{OT}}$:
$\mathcal{L}_{\mathrm{FM}}$ is optimized on adjacent pairs $(t_a,t_b)\in\mathcal{E}_{\mathrm{tr}}$ every iteration,
while $\mathcal{L}_{\mathrm{OT}}$ and $\mathcal{L}_{\mathrm{dyn}}$ are evaluated and backpropagated only once every $K_{\mathrm{OT}}$ iterations.

\begin{algorithm}[t]
\caption{\model}
\label{alg:cellfm_short}
\DontPrintSemicolon
\SetKwInput{KwIn}{Input}
\SetKwInput{KwOut}{Output}

\KwIn{$T_{\mathrm{tr}}$; snapshots $\{X_t\}_{t\in T_{\mathrm{tr}}}$;
$\varepsilon$, $E_{\mathrm{warm}}$, $K_{OT}$, and loss weights in \Eq{eq:full-objective};
$Enc_\phi$, $Dec_\theta$, $v_\psi^{\mathrm{f}}$, $v_\psi^{\mathrm{b}}$.}
\KwOut{Trained parameters $(\phi,\theta,\psi)$.}

\tcp{Phase I: VAE pretraining}
Optimize $(\phi,\theta)$ on $\mathrm{CONCAT}(\{X_t\})$ to minimize $\mathcal{L}_{\mathrm{VAE}}${\scriptsize(\Eq{eq:vae-loss})}\;

\tcp{Phase II: dynamics learning}
\For(\tcp*[f]{each training step $s$}){$s=1,2,\dots$}{
  Sample adjacent $(t_a,t_b)$ and mini-batches from $X_{t_a},X_{t_b}$; encode to $Z_a,Z_b$\;
  Compute $\mathcal{L}_{\mathrm{VAE}}$ on the sampled mini-batches\;

  \uIf{$s \le E_{\mathrm{warm}}$}{
    $C \leftarrow C^{\mathrm{euc}}(Z_a,Z_b)$\ {\scriptsize(\Eq{eq:cost-euc})}\;
  }\Else{
    Compute one-step Euler predictions $\tilde Z_b,\tilde Z_a$ using $v_\psi^{\mathrm{f}},v_\psi^{\mathrm{b}}${\scriptsize(\Eq{eq:euler-pred})}\;
    $C \leftarrow C^{\mathrm{bi}}(Z_a,Z_b,\tilde Z_a,\tilde Z_b)${\scriptsize(\Eq{eq:cost-bi})}
  }
  Compute entropic OT coupling $\pi \leftarrow \mathrm{Sinkhorn}(C)$ {\scriptsize(\Eq{eq:sinkhorn})}\;

  Compute local $\mathcal{L}_{\mathrm{FM}}$ (Top-$K$ OT approximation){\scriptsize(\Eqs{eq:fm-fwd}{eq:fm-topk-fwd})}\;
  $L \leftarrow \mathcal{L}_{\mathrm{VAE}} + \lambda_{\mathrm{FM}}\mathcal{L}_{\mathrm{FM}}${\scriptsize(\Eq{eq:full-objective})}\;
  \If{$s \bmod K_{OT}=0$}{
    Compute $\mathcal{L}_{\mathrm{OT}},\mathcal{L}_{\mathrm{dyn}}${\scriptsize(\Eq{eq:ot-loss}, \Eq{eq:dyn-reg})}\;
    $L \leftarrow L + \lambda_{\mathrm{OT}}\mathcal{L}_{\mathrm{OT}}+\lambda_{\mathrm{dyn}}\mathcal{L}_{\mathrm{dyn}}${\scriptsize(\Eq{eq:full-objective})}\;
  }
  Update $(\phi,\theta,\psi)$ by minimizing $L$\;
  \If{converged}{break\;}
}
\Return{$(\phi,\theta,\psi)$}\;
\end{algorithm}

\subsection{Theoretical Analysis}
\label{sec:theory-overview}
We provide a theoretical analysis for our OT-conditioned flow matching dynamics learner in the latent space.
Specifically, we establish well-posedness of the induced ODE flow under standard regularity assumptions,
prove Wasserstein stability bounds that translate velocity-field regression error into endpoint distribution error,
characterize expressivity and finite-sample generalization of neural velocity fields via standard approximation and complexity arguments,
and provide a nonconvex optimization guarantee showing stochastic gradient methods converge to stationary points.
Full statements and proofs are provided in Appendix.

%% file: chapter/3_exp.tex
\section{Experiment}
\subsection{Experimental Setup}
\label{subsec:datasets}
\paragraph{Datasets}
We evaluate on three widely used time-series scRNA-seq benchmarks that cover diverse biological regimes, from rapid embryogenesis to long-horizon reprogramming:
\textbf{ZB} (zebrafish embryo)~\cite{farrell2018single}, \textbf{DR} (drosophila)~\cite{calderon2022continuum}, and \textbf{SC} (Schiebinger2019)~\cite{schiebinger2019optimal}.
These datasets are challenging due to strong distribution shift across timepoints, asynchronous sampling, and the absence of cell-wise correspondences. \\

\paragraph{Baselines}
We compare with representative methods spanning the major paradigms for time-series scRNA-seq: (1) Optimal-transport-based population matching: \textbf{WOT}. (2) Continuous-time latent dynamics models: \textbf{scNODE} and \textbf{PRESCIENT}. (3) Flow-matching / OT-regularized continuous dynamics: \textbf{VGFM}. (4) Non-learning sanity check: \textbf{Naive} replicates the last observed timepoint for future prediction.

\paragraph{Evaluation metrics}
At each held-out timepoint $t$, we compare the predicted cell set $\widehat{X}(t) \in \mathbb{R}^{n_2 \times G}$ against the true cell set $X(t) \in \mathbb{R}^{n_1 \times G}$ using:
(i) \textbf{2-Wasserstein distance} (approximated by Sinkhorn), and
(ii) \textbf{average pairwise $\ell_2$ distance}
\begin{equation}
\ell_2(X(t), \widehat{X}(t)) \;=\; \frac{1}{n_1 n_2}\sum_{i=1}^{n_1}\sum_{j=1}^{n_2}\left\|X_i(t)-\widehat{X}_j(t)\right\|_2.
\label{eq:l2_metric}
\end{equation}
We report metrics per held-out timepoint and summarize by averaging across all held-out timepoints.

\paragraph{Implemention Details}
For each dataset, we library-size normalize counts, apply $\log(1+x)$, and retain the top 2{,}000 HVGs. 
All methods use the same preprocessed matrices for fair comparison.
We use a VAE with latent dimension 50 and a continuous-time velocity field parameterized by a residual network (hidden size 256) conditioned on sinusoidal time embeddings (dimension 64). We integrate the learned ODE with RK4 (step size 0.1). 
Training consists of (1) VAE pretraining and (2) flow-matching dynamics learning. 
To stabilize minibatch couplings in early training, we warm up with Euclidean-cost couplings and then switch to the full bidirectional fused-cost couplings. 
We periodically apply OT- and dynamics-related regularizers during training.
We compute Sinkhorn-based OT objectives with blur=0.05 and scaling=0.5.
OT-only matching methods are only applicable to interpolation, and we exclude baselines that require training a separate model for each held-out timepoint in the multi-holdout extrapolation setting.

\begin{table}[t]
  \centering
  \caption{Summary of datasets used in our experiments. We follow a consistent preprocessing protocol and select 2{,}000 HVGs for all datasets. The timepoint index for interpolation and extrapolation tasks in each dataset starts with 0.}
  \vspace{-2mm}
  \resizebox{\linewidth}{!}{%
  \label{tab:dataset_split}
  \begin{tabular}{cccccc}
    \toprule
    \multirow{2}{*}{\textbf{Data}} & 
    \multirow{2}{*}{\textbf{\# Cells}} & 
    \multirow{2}{*}{\textbf{\# Genes(HVG)}} & 
    \multirow{2}{*}{\textbf{\# Timepoints}} & 
    \multicolumn{2}{c}{\textbf{Testing Timepoints}} \\
    \cmidrule(l){5-6}
    & & & & Interpolating & Extrapolating \\
    \midrule
    \textbf{ZB} & 38,731  & 2,000 & 12 & 4, 6, 8    & 10, 11 \\
    \textbf{DR} & 27,386  & 2,000 & 11 & 4, 6, 8    & 8, 9, 10 \\
    \textbf{SC} & 236,285 & 2,000 & 19 & 5, 10, 15  & 16, 17, 18 \\
    \bottomrule
  \end{tabular}}
\end{table}

\begin{table*}[t]
  \centering
  \caption{Wasserstein distance (denoted as $W$) and $\ell_2$ distance between left out timepoints and model predictions.
  For each dataset block, the first row indicates the left-out timepoints for interpolation and extrapolation.
  Bold numbers denote the best prediction. ``NA'' indicates the model is inapplicable. ``-'' indicates the timepoint is not present for that dataset.}
  \label{tab:all_wass_l2}
  \small
  \setlength{\tabcolsep}{5pt}
  \renewcommand{\arraystretch}{1.15}
  \begin{adjustbox}{max width=\textwidth}
    \begin{tabular}{ll *{6}{cc}}
      \toprule
      \multirow{2}{*}{\textbf{Dataset}} & \multirow{2}{*}{\textbf{Model}} &
        \multicolumn{6}{c}{\textbf{Interpolation}} &
        \multicolumn{6}{c}{\textbf{Extrapolation}} \\
      \cmidrule(lr){3-8}\cmidrule(lr){9-14}
      & & $W$ & $\ell_2$ & $W$ & $\ell_2$ & $W$ & $\ell_2$
        & $W$ & $\ell_2$ & $W$ & $\ell_2$ & $W$ & $\ell_2$ \\
      \midrule

      \multirow{7}{*}{\textbf{ZB}}
      & \textit{Left-out $t$}
      & \multicolumn{2}{c}{$t=4$}
      & \multicolumn{2}{c}{$t=6$}
      & \multicolumn{2}{c}{$t=8$}
      & \multicolumn{2}{c}{$t=10$}
      & \multicolumn{2}{c}{$t=11$}
      & \multicolumn{2}{c}{-} \\
      \cmidrule(lr){3-4}\cmidrule(lr){5-6}\cmidrule(lr){7-8}\cmidrule(lr){9-10}\cmidrule(lr){11-12}\cmidrule(lr){13-14}
      & \textbf{\model}
      & \textbf{426.58} & \textbf{31.52}
      & \textbf{375.51} & \textbf{30.70}
      & \textbf{458.99} & \textbf{33.92}
      & \textbf{599.98} & \textbf{37.93}
      & \textbf{653.19} & \textbf{39.38}
      & - & - \\
      & VGFM
      & 720.27 & 41.04
      & 688.72 & 40.24
      & 764.56 & 42.76
      & 890.42 & 45.85
      & 1056.08 & 49.59
      & - & - \\
      & scNODE
      & 434.28 & 32.35
      & 390.50 & 31.15
      & 468.41 & 34.57
      & 607.87 & 38.78
      & 685.47 & 40.78
      & - & - \\
      & PRESCIENT
      & 830.88 & 43.03
      & 662.15 & 39.17
      & 669.46 & 40.32
      & 1063.63 & 49.09
      & 1112.12 & 49.97
      & - & - \\
      & WOT
      & 575.11 & 38.17
      & 492.84 & 37.01
      & 565.70 & 40.64
      & NA & NA
      & NA & NA
      & - & - \\
      & Naive
      & 775.63 & 43.88
      & 695.72 & 42.28
      & 720.77 & 44.68
      & 887.69 & 49.06
      & 943.98 & 50.08
      & - & - \\

      \midrule

      \multirow{7}{*}{\textbf{DR}}
      & \textit{Left-out $t$}
      & \multicolumn{2}{c}{$t=4$}
      & \multicolumn{2}{c}{$t=6$}
      & \multicolumn{2}{c}{$t=8$}
      & \multicolumn{2}{c}{$t=8$}
      & \multicolumn{2}{c}{$t=9$}
      & \multicolumn{2}{c}{$t=10$} \\
      \cmidrule(lr){3-4}\cmidrule(lr){5-6}\cmidrule(lr){7-8}\cmidrule(lr){9-10}\cmidrule(lr){11-12}\cmidrule(lr){13-14}
      & \textbf{\model}
      & \textbf{352.94} & \textbf{28.95}
      & \textbf{403.01} & \textbf{30.95}
      & 493.99 & \textbf{34.61}
      & \textbf{527.01} & \textbf{36.07}
      & \textbf{529.69} & \textbf{35.77}
      & \textbf{682.66} & \textbf{39.74} \\
      & VGFM
      & 509.33 & 34.62
      & 568.96 & 36.32
      & 701.13 & 39.64
      & 766.84 & 43.58
      & 934.79 & 48.18
      & 1431.02 & 58.41 \\
      & scNODE
      & 354.10 & 29.25
      & 412.94 & 31.76
      & \textbf{486.53} & 34.99
      & 547.91 & 36.93
      & 554.60 & 37.21
      & 691.96 & 40.47 \\
      & PRESCIENT
      & 447.31 & 32.09
      & 474.97 & 33.22
      & 519.20 & 35.24
      & 583.07 & 37.20
      & 536.83 & 36.16
      & 692.53 & 39.80 \\
      & WOT
      & 426.70 & 34.35
      & 508.96 & 37.06
      & 592.67 & 40.52
      & NA & NA
      & NA & NA
      & NA & NA \\
      & Naive
      & 529.27 & 38.54
      & 607.85 & 41.14
      & 760.64 & 45.58
      & 824.57 & 47.40
      & 824.57 & 46.90
      & 969.15 & 49.50 \\

      \midrule

      \multirow{7}{*}{\textbf{SC}}
      & \textit{Left-out $t$}
      & \multicolumn{2}{c}{$t=5$}
      & \multicolumn{2}{c}{$t=10$}
      & \multicolumn{2}{c}{$t=15$}
      & \multicolumn{2}{c}{$t=16$}
      & \multicolumn{2}{c}{$t=17$}
      & \multicolumn{2}{c}{$t=18$} \\
      \cmidrule(lr){3-4}\cmidrule(lr){5-6}\cmidrule(lr){7-8}\cmidrule(lr){9-10}\cmidrule(lr){11-12}\cmidrule(lr){13-14}
      & \textbf{\model}
      & \textbf{56.72} & \textbf{12.05}
      & \textbf{125.34} & \textbf{19.57}
      & \textbf{101.52} & \textbf{19.68}
      & \textbf{112.14} & \textbf{19.44}
      & \textbf{112.76} & \textbf{19.32}
      & \textbf{121.42} & 19.62 \\
      & VGFM
      & 110.43 & 16.08
      & 203.09 & 21.97
      & 197.53 & 22.84
      & 218.92 & 22.49
      & 224.14 & 22.69
      & 225.65 & 22.68 \\
      & scNODE
      & 61.15 & 12.46
      & 139.39 & 20.09
      & 116.07 & 20.63
      & 122.58 & 19.88
      & 123.66 & 19.73
      & 126.45 & \textbf{19.18} \\
      & PRESCIENT
      & 106.14 & 16.37
      & 129.83 & 18.60
      & 143.94 & 20.91
      & 134.32 & 20.38
      & 142.13 & 20.99
      & 140.70 & 20.60 \\
      & WOT
      & 71.36 & 14.68
      & 115.90 & 18.79
      & 143.06 & 23.66
      & NA & NA
      & NA & NA
      & NA & NA \\
      & Naive
      & 83.53 & 16.24
      & 148.45 & 21.41
      & 192.04 & 26.74
      & 161.35 & 25.13
      & 169.72 & 25.89
      & 167.25 & 26.45 \\

      \bottomrule
    \end{tabular}
  \end{adjustbox}
\end{table*}

\subsection{Unmeasured Timepoints Prediction}
\begin{figure}[htbp]%
    \centering { \includegraphics[width=\linewidth]{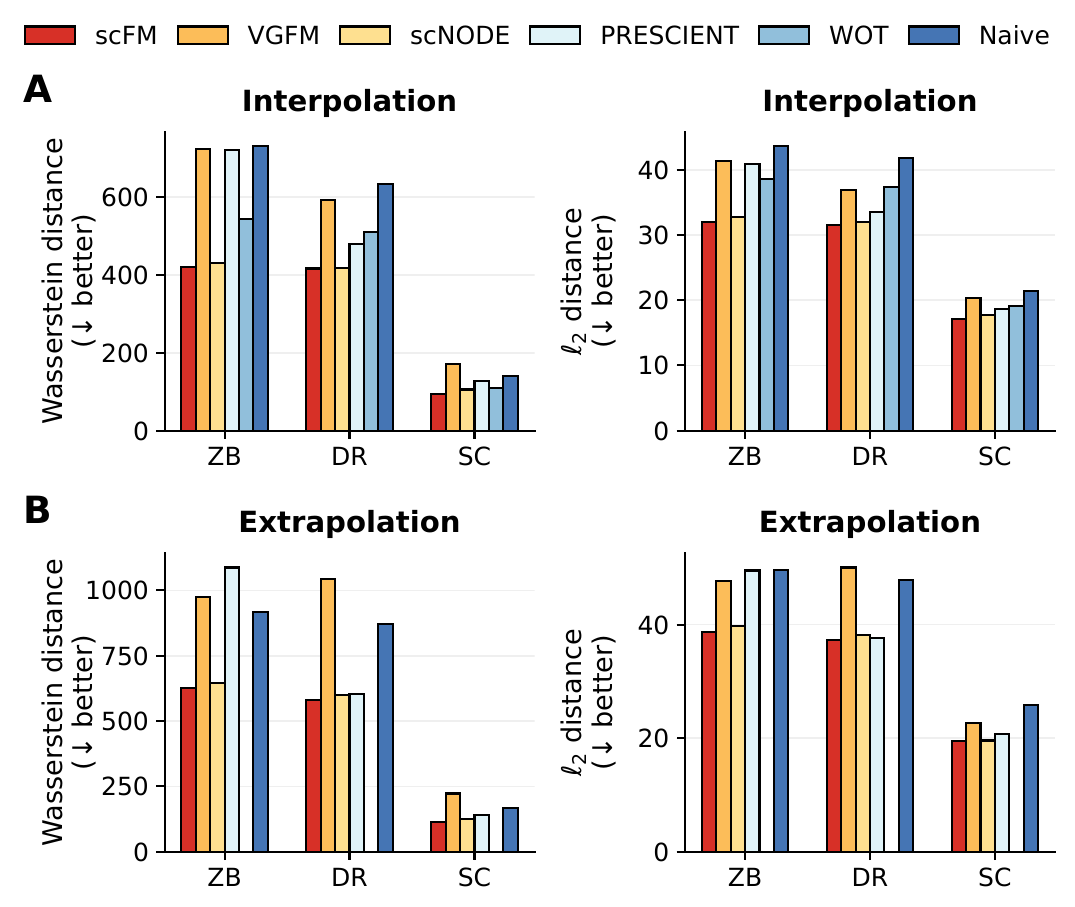} }
    \vspace{-6mm}
    \caption{Averaged Wasserstein and $\ell_2$ distance between true and predicted expressions on unmeasured timepoints.} 
    \label{fig:distance}
    \vspace{-2mm}
\end{figure}

We adopt the common \emph{multi-holdout} protocol, where several timepoints are held out during training and are predicted from remaining observed snapshots.
Interpolation targets missing \emph{intermediate} timepoints bracketed by observed neighbors, whereas extrapolation targets \emph{future} timepoints beyond the latest training snapshot.
The exact splits for ZB, DR, and SC are summarized in Table~\ref{tab:dataset_split}.
Notably, in all settings we train \emph{a single model per dataset} and query it at arbitrary held-out times, which directly tests temporal generalization under distribution shift.
Fig.~\ref{fig:distance} reports the performance on unmeasured timepoints using both wasserstein distance and $\ell_2$ distance.
Across all three datasets, \model{} consistently achieves lower distances than prior continuous-time latent-dynamics baselines, indicating more accurate snapshot distribution prediction at held-out times.
Importantly, the ranking induced by Wasserstein and $\ell_2$ is largely consistent, suggesting that \model{} improves both global distribution matching and pointwise feature-space proximity rather than optimizing a single metric in isolation.

For \textbf{interpolation}, OT-based matching is applicable and often serves as a strong reference.
In this regime, \model{} remains competitive with the strongest OT-style baseline while outperforming neural ODE baselines, demonstrating that learning continuous-time dynamics does not come at the cost of intermediate-time reconstruction quality.
For \textbf{extrapolation}, OT-only methods are not applicable (NA) since they require bracketing observed snapshots.
In contrast, \model{} maintains robust long-horizon predictive accuracy and exhibits markedly smaller degradation as the prediction horizon increases, highlighting stronger temporal extrapolation and better handling of progressive distribution shift.

\subsection{Trajectory Recovery}
\begin{figure}[htbp]
    \centering
    \includegraphics[width=\linewidth]{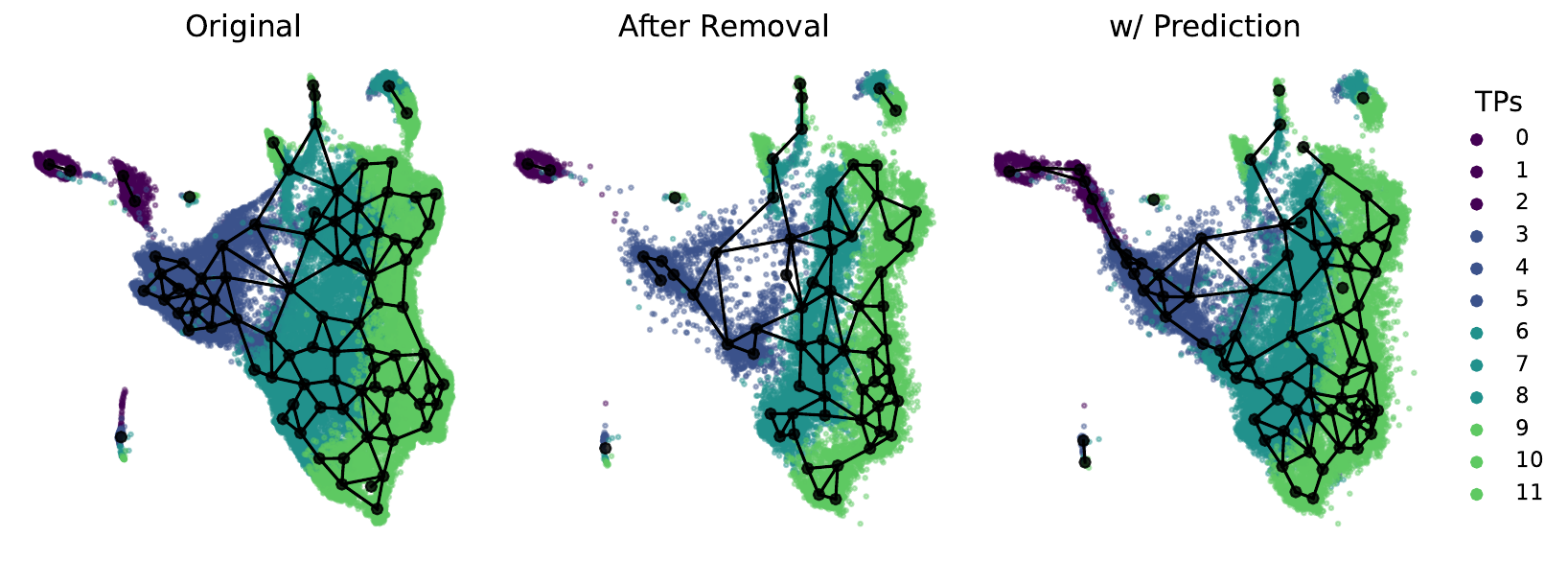}
    \vspace{-5mm}
    \caption{\textbf{Trajectory recovery on Zebrafish under the remove-recovery setting.}
    Left: PAGA graph built from the full dataset. 
    Middle: graph after removing target timepoints.
    Right: graph augmented with \model-predicted cells at the removed timepoints. 
    }
    \label{fig:recovery_zb}
\end{figure}

We further evaluate whether model-predicted cells can \emph{recover} trajectory structure when intermediate timepoints are missing. Following the standard remove-recovery protocol, we first build the PAGA connectivity graph using all observed timepoints. We then remove a subset of timepoints and rebuild the graph on the remaining data, which often introduces discontinuities and topology distortions due to missing transitional populations. Finally, we add model-predicted cells at the removed timepoints and rebuild the PAGA graph. We quantify divergence using the Ipsen–Mikhailov (IM) distance to the full graph; lower IM indicates closer agreement.
Fig.~\ref{fig:recovery_zb} shows an example on ZB: compared to the reduced graph, adding predicted cells restores transitional regions and yields a connectivity pattern closer to the reference. Across datasets, \model{} consistently reduces IM distance, indicating better preservation of global connectivity after removing intermediate timepoints.

\subsection{Visualization}
\begin{figure}[htbp]%
    \centering { \includegraphics[width=\linewidth]{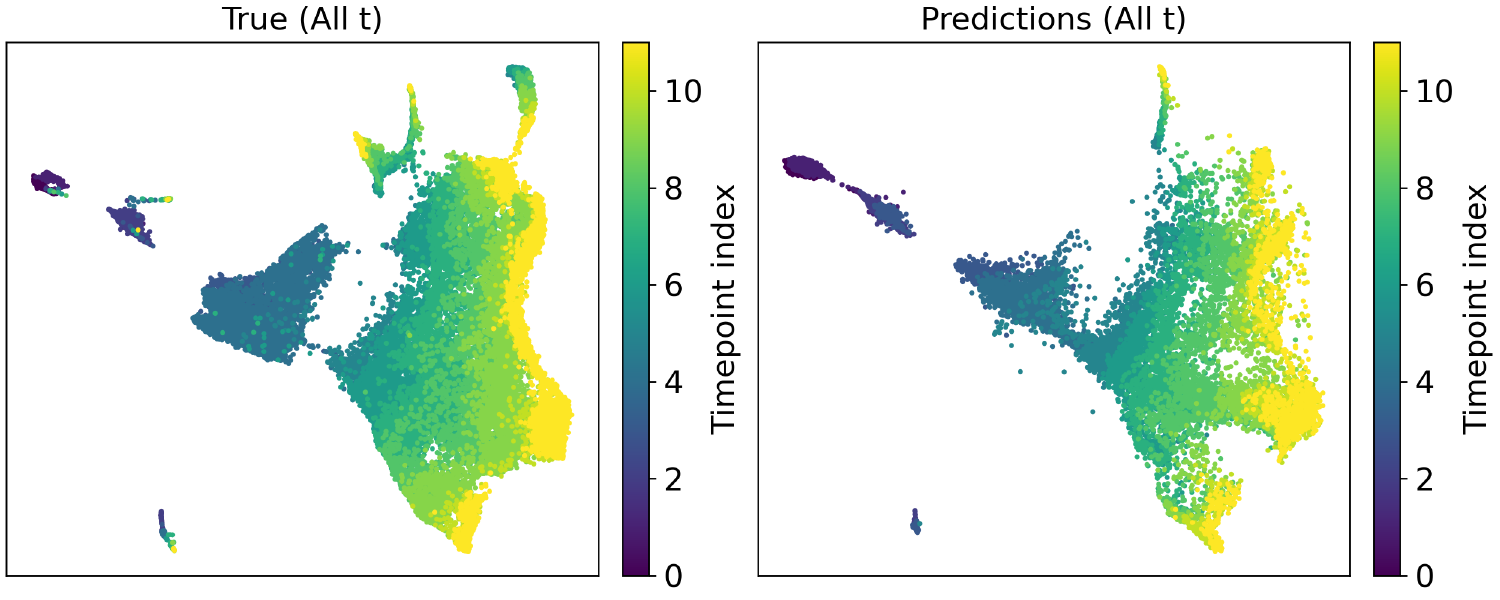} }
    \vspace{-5mm}
    \caption{2D UMAP visualization of true and predicted expression of interpolation task on ZB dataset} 
    \label{fig:interpolation}
    \vspace{-2mm}
\end{figure}

To provide an intuitive, distribution-level comparison, we further project both the ground-truth cells and the model predictions into a shared two-dimensional UMAP embedding.
We focus on the interpolation setting on the ZB dataset and color each cell by its timepoint index to highlight temporal progression.
As shown in Figure~\ref{fig:interpolation}, the predicted trajectories largely preserve the global manifold geometry and exhibit a temporal gradient consistent with the ground truth, indicating that the model captures the underlying developmental dynamics in the interpolated regime.

%% file: chapter/4_conclusion.tex
\section{Conclusion}
This paper proposes \model, a latent generative framework based on coupling-conditioned flow matching. First, we compute entropically regularized OT couplings between adjacent snapshots and use them to construct soft, weighted flow-matching targets for learning time-dependent velocity fields. Second, we learn bidirectional velocity fields and leverage their consistency to refine couplings and improve temporal coherence under sparse supervision. Third, we introduce distribution-level alignment and latent dynamic regularization to anchor long rollouts and mitigate drift. Experiments on real-world time-series scRNA-seq datasets show that \model~consistently improves distributional prediction performance for both temporal interpolation and extrapolation. Moreover, \model~yields more accurate trajectory reconstruction and temporally coherent visualizations where intermediate time points are absent, indicating a more faithful recovery of underlying temporal gene expression dynamics.